\documentclass[12pt]{article}

\usepackage[french]{babel}

\usepackage[letterpaper,top=2cm,bottom=2cm,left=3cm,right=3cm,marginparwidth=1.75cm]{geometry}

\usepackage{amsmath}
\usepackage{graphicx}
\usepackage[colorlinks=true, allcolors=blue]{hyperref}
\usepackage{ dsfont }
\usepackage{stmaryrd}
\usepackage{amsthm}

\usepackage{booktabs}
\usepackage{longtable}

\usepackage{caption}
\usepackage{subcaption}

\usepackage{natbib}
\title{Régression ordinale pour prédictions}

\DeclareMathOperator*{\argmax}{arg\,max}

\begin{document}

\theoremstyle{preuve}
\newtheorem{preuve}{Proposition}

\theoremstyle{prop}
\newtheorem{prop}{Proposition}

\theoremstyle{definition}
\newtheorem{definition}{Definition}[section]

\renewcommand\refname{Bibliographie}

\begin{center}
{\Large
	{\sc Prédiction optimale pour un modèle ordinal à covariables fonctionnelles}
}
\bigskip

 \underline{Simón Weinberger} $^{1}$ \& Jairo Cugliari $^{2}$ \& Aurélie Le Cain$^{3}$
\bigskip

{\it
$^{1}$ Laboratoire ERIC, EssilorLuxottica, France, weinbes@essilor.fr\\
$^{2}$ Laboratoire ERIC, France, Jairo.Cugliari@univ-lyon2.fr \\
$^{3}$ EssilorLuxottica, France, lecaina@essilor.fr
}
\end{center}
\bigskip


{\bf R\'esum\'e.} On présente un cadre formel prédictif pour le modèle ordinal cumulatif. Dans ce cadre, on donne la forme explicite d'une prédiction optimale qui minimise la moyenne de l'erreur en valeur absolue. De plus, on réduit un modèle de régression ordinale à covariables fonctionnelles, à un modèle ordinal classique, à covariables scalaires. On illustre les méthodes proposées et on essaye d'appliquer ce modèle sur des données collectées par EssilorLuxottica dans le cadre du développement d'un algorithme pour le contrôle de la teinte d'une monture à verre actif.

{\bf Mots-cl\'es.} Classification supervisée et non supervisée, Apprentissage statistique, Données fonctionnelles.

\medskip

{\bf Abstract.} We present a prediction framework for ordinal models: we introduce optimal predictions using loss functions and give the explicit form of the Least-Absolute-Deviation prediction for these models. Then, we reformulate an ordinal model with functional covariates to a classic ordinal model with mutliple scalar covariates. We illustrate all the proposed methods and try to apply these to a dataset collected by EssilorLuxottica for the developpement of a control algorithm for the shade of connected glasses.


\bigskip\bigskip

\section{Introduction}



Le modèle de régression ordinal, avec variables explicatives fonctionnelles, est une extension du modèle ordinal standard qui permet de modéliser la relation entre une variable réponse ordinale et des variables explicatives fonctionnelles. Il s'agit de décrire, expliquer et prédire une variable ordinale à partir d'une donnée fonctionnelle \cite{fda_ramsay}. On considère une application industrielle, où une variable ordinale doit être prédite à partir des données échantillonnées par un capteur intégré dans un dispositif mobile. 

Grâce aux technologies portables (\textit{wearables}) il est de plus en plus commun d'avoir des mesures successives d'un phénomène au cours du temps. On peut citer des mesures de la température à chaque minute \cite{fda_ramsay}, le nombre d'œufs pondus par une mouche chaque jour \cite{mouches} ou la résistance d'une pâte à chaque seconde, lorsqu'on la pétrit \cite{cookies}. 

On s'intéresse au développement de lunettes avec un verre actif e-chromic dont la teinte peut être modulée grâce à des capteurs électroniques. 
Concrètement, on a quatre classes de teintes distinctes, plus ou moins foncées, et on essayera de prédire la classe que l'utilisateur choisira, en fonction de la luminosité ambiante dans une fenêtre temporelle. On s'intéresse donc à une donnée ordinale, la teinte du verre, expliquée par une donnée fonctionnelle, la valeur de la lumière ambiante mesurée à une certaine fréquence. 
De plus, la prise de décision doit se faire économiquement vu qu'on travaille avec un système embarqué: on a des contraintes liées à l'utilisation de la batterie et on a un espace mémoire limité.

Deux références sont d'intérêt. D'une part, \cite{mouches} traite de façon générale l'inférence pour des modèles linéaires généralisés à covariables fonctionnelles, dont le modèle ordinal est un cas particulier. D'autre part, \cite{regOrd_fonctionelle} construit explicitement le modèle ordinal cumulatif à covariables fonctionnelles. Dans les deux cas, l'intérêt est centré dans la construction du modèle, puis dans son estimation et inférence.

Dans ce document, on reprendra le travail effectué dans \cite{regOrd_fonctionelle} mais on adoptera un point de vue prédictif. On s'intéressera ainsi à la forme des prédictions \og optimales \fg{}, c'est à dire qui minimisent, en moyenne, l'erreur évalué en utilisant une fonction de perte donnée, c'est le point de vue adopté dans \cite{stat_learning}. De plus on proposera de rajouter une pénalisation LASSO à l'estimation par maximum de vraisemblance du modèle proposé dans \cite{regOrd_fonctionelle}.

Le reste de ce document est organisé de la façon suivante.

Dans la section 2, on présente le modèle ordinal, on pose le cadre nécessaire pour faire des prédictions et on présente la forme explicite d'une prédiction qui est optimale vis à vis d'une fonction de perte dans un sens précis.
Dans la section 3, on s'intéresse à un modèle ordinal à covariables fonctionnelles, ensuite on montre qu'on peut réduire ce modèle à une régression ordinale classique avec plusieurs covariables scalaires et on évoquera la possibilité d'utiliser une pénalisation LASSO sur ce modèle réduit.
Dans la section 4, on applique ce modèle sur le jeu de données \og Kneading \fg{} \cite{regOrd_fonctionelle}. Ensuite on utilise ce même modèle sur des données collectées par EssilorLuxottica, en implémentant une pénalisation LASSO.

\section{Modèle ordinal}

Le modèle ordinal est un cas particulier du modèle linéaire généralisé où la variable expliquée, $Y$, est discrète avec $K$ valeurs possibles, avec $K$ un entier naturel connu. On supposera en plus qu'il existe une relation d'ordre entre les valeurs que $Y$ peut prendre. On dit alors que $Y$ est une variable ordinale.

Sans perdre de généralité, on peut supposer que $Y$ est à valeurs dans $\mathcal{S} = \{ 1, \ldots, K\}$. Le modèle ordinal modélise alors les probabilités $(\mathds{P}(Y\leq j))_{j\in\mathcal{S}}$ avec une fonction de lien et une relation entre les variables explicatives et ces probabilités $(\mathds{P}(Y\leq j))_{j\in\mathcal{S}}$. Habituellement on suppose que cette relation est linéaire, mais dans cette section on adopte un point légèrement plus général, on permet que cette relation soit non linéaire. 

\subsection{Cadre et définitions}


Soit $F$ une fonction de répartition d'une variable aléatoire continue avec $F$ strictement croissante, soit $\mathcal{X}$ un espace de covariables et soit $g$ une fonction de $\mathcal{X}$ vers $\mathds{R}$. On considère le modèle:
\begin{align}
    F^{-1}(\mathds{P}(Y \leq j)) =  \tau_j - g(X)\quad ; \quad j\in\mathcal{S} \label{eq:ordinal_mod}
\end{align}
où $X\in \mathcal{X}$ est une covariable et  $(\tau_i)_{i\in\llbracket0,K\rrbracket}$ sont des seuils ordonnés:
$$ -\infty = \tau_0 < \tau_1 < \tau_2 < \ldots <\tau_{K-2} < \tau_{K-1} < \tau_K = \infty . $$
Le choix de ce modèle est justifié par l'utilisation d'une variable latente $Y^*$, à valeur dans $\mathds{R}$. En effet, soit $\epsilon$ une variable aléatoire qui admet $F$ comme fonction de répartition, avec $F$ strictement croissante. On supposera que la variable aléatoire $Y$ est expliquée par une variable aléatoire latente $Y^*$ de la façon suivante:
\begin{align}
    &Y^* = g(X) + \epsilon \label{eq:latent_1}, \\ 
    &Y = j \quad  \Leftrightarrow \quad \tau_{j-1} < Y^* \leq \tau_j \quad ; \quad j \in \mathcal{S}\label{eq:latent_2} .
\end{align}
Notons qu'avec les lignes (\ref{eq:latent_1}) et (\ref{eq:latent_2}), on retrouve la ligne (\ref{eq:ordinal_mod}).

Dans ce document, on s'intéresse à deux espaces de covariables différents: $\mathds{R}^d$ et $L^2([0,T])$, l'espace de fonctions d'un intervalle $[0,T]$ à valeurs dans $\mathds{R}$, de carré intégrable. Ces deux espaces sont munis de produits scalaires usuels, que l'on notera respectivement $\langle . , . \rangle_{\mathds{R}^d}$ et $\langle . , . \rangle_{L^2([0,T])}$.

Soit $\boldsymbol{x}=(x_1,\ldots,x_d)$ et $\boldsymbol{b}=(b_1,\ldots, b_d)$ deux points de $\mathds{R}^d$ et soit $X$ et  $\beta$ deux points de $L^2([0,T])$. Les produits scalaires de chaque espace sont définis comme il suit:
$$ \langle \boldsymbol{x} ,\boldsymbol{b} \rangle_{\mathds{R}^d} = \sum_{i=1}^d x_i b_i \quad ; \quad \langle X, \beta \rangle_{L^2([0,T])} = \int_{[0,T]} X(s)\beta(s)ds . $$
Posons $logit(p)=log(\frac{p}{1-p})$, notons que $logit$ est l'inverse de la fonction de répartition de la loi logistique standard.

Le modèle ordinal cumulatif classique présenté dans \cite{agresti_categorical} est définit par le modèle suivant:
$$ logit(\mathds{P}(Y \leq j)) =  \tau_j - \langle \boldsymbol{x} ,\boldsymbol{b} \rangle_{\mathds{R}^d} \quad ; \quad j\in\mathcal{S} .$$
Les paramètres de ce modèle sont les seuils $(\tau_j)_{j\in\mathcal{S}}$ et le vecteur $\boldsymbol{b}\in\mathds{R}^d$.

Le modèle ordinal cumulatif à covariables fonctionnelles présenté dans \cite{regOrd_fonctionelle} est définit par:
$$ logit(\mathds{P}(Y \leq j)) =  \tau_j - \langle X ,\beta \rangle_{L^2([0,T])} \quad ; \quad j\in\mathcal{S} . $$
Les paramètres de ce modèle sont les seuils $(\tau_j)_{j\in\mathcal{S}}$ et la fonction $\beta\in L^2([0,T])$.

Si l'on note $\pi_j = \mathds{P}(Y=j)$, alors on a:
\begin{align*}
    \pi_j = \mathds{P}(Y = j) &= \mathds{P}(\tau_{j-1} < Y^* \leq \tau_j) = \mathds{P}(X \cdot \beta + \epsilon \leq \tau_j) - \mathds{P}(X \cdot \beta + \epsilon \leq \tau_{j-1}).
\end{align*}
%

Notons que jusqu'ici on a défini tout ce qui est nécessaire pour faire de l'inférence avec ce modèle. Hors, on ne peux pas encore faire de la prédiction avec le cadre posé jusque ici. Dans la sous section suivante on illustre pourquoi et on reprend le cadre de théorie de la décision utilisé en \cite{stat_learning}.
\subsection{Prédiction}

Étudions de façon formelle les \og règles de décision \fg{} qu'on peut obtenir avec ce modèle, ce sont des fonctions qui permettraient de décider quelle classe il faudrait prédire à partir d'une covariable $X\in\mathcal{X}$ donnée. 

Concrètement, une règle de décision est une fonction $r$ de l'espace des covariables $\mathcal{X}$ vers l'espace des états $\mathcal{S}$:
    $$
    r: \mathcal{X} \to \mathcal{S}.
    $$
    Une fonction de coût, $C$, est une fonction:
    $$
    C: \mathcal{X}\times\mathcal{X} \to \mathds{R}^+.
    $$
    Soit $Y$ une variable aléatoire à valeurs dans $\mathcal{S}$, une règle de décision optimale pour une fonction de perte $C$ est une règle de décision $r_{opt}$ vérifiant:
    $$ 
    \forall x \in \mathcal{X}, \quad r_{opt}(x) = \inf_{\hat{y}\in \mathcal{S}} {\mathds{E}(C(\hat{y},Y) | X = x)}. 
    $$

Par exemple, on peut considérer les deux règles de décision suivantes:
\begin{align*}
    &r_{(1)}(X) = \argmax_{y\in \mathcal{S}} \pi_{y}(X),\\
    &r_{(2)}(X) = j \quad \Leftrightarrow \quad \tau_{j-1} < g(X) \leq \tau_j \quad j \in \mathcal{S}.
\end{align*}
Pour une covariable fixe $X$, $r_{(1)}(X)$ prédit la classe la plus probable et $r_{(2)}(X)$ prédit la classe qu'on aurait eue en remplaçant $Y^*$ par $g(X)$ dans (\ref{eq:latent_2}). Elles sont toutes les deux des prédictions \og raisonnables \fg{} et elles sont en réalité optimales pour des fonctions de coût différentes: $r_{(1)}$ est optimale pour $C(\hat{y},y) = \mathds{1}_{\hat{y} \neq y}$ et $r_{(2)}$ est optimale pour $C(\hat{y},y) = | \hat{y} - y |$.

Dans la suite on fera référence à $r_{(2)}$, par prédiction \og LAD \fg{}, comme \og Least Absolute Difference \fg{}.

Notons que si l'on veut faire une prédiction optimale vis-à-vis de la différence en valeur absolue ($C(\hat{y},y) = | \hat{y} - y |$), on peut s'en passer des calculs des probabilités $\pi(X)$. Ceci est intéressant en termes de puissance de calcul, on a juste besoin de calculer un produit scalaire et le comparer à des seuils. Par contre si on veut une prédiction optimale vis-à-vis de $C(\hat{y},y) = \mathds{1}_{\hat{y}!=y}$, il faudra calculer $\pi(X)$, vu qu'on ne sait pas donner de forme explicite de $r_{(1)}(X)$ en fonction de $X$.

\section{Modèle ordinal à covariables fonctionnelles}

Dans \cite{regOrd_fonctionelle} il a été proposé d'utiliser un modèle ordinal à covariables fonctionnelles. On reprend ce modèle et on montre que sous les hypothèses de ce même article, on peut réduire le modèle à un modèle ordinal classique. On évoquera en plus la possibilité de rajouter un terme de pénalisation LASSO sur ce modèle à covariables fonctionnelles.

\subsection{Modele FOLR (\og Functional Ordinal Logistic Regression \fg{})}

On reprend le modèle introduit dans \cite{regOrd_fonctionelle}:
\begin{align*}
        &Y^* = \langle X ,\beta \rangle_{L^2([0,T])} + \epsilon \stepcounter{equation}\tag{\theequation}\label{eq:FOLR},\\
        &Y = j \quad \Leftrightarrow \quad \tau_{j-1} < Y^*  \leq \tau_j.
\end{align*}
Où $X$ et $\beta$ sont des fonctions dans $L^2([0,T])$. $X$ est une covariable fonctionelle et $\beta$ est une fonction inconnue. On suppose que $\epsilon$ suit une distribution logistique standard.

\subsection{Réduction du modèle FOLR}

Dans \cite{regOrd_fonctionelle}, on fait une hypothèse sur la forme des fonctions $X$ et $\beta$:
$$
X(t) = \sum_{r=1}^R a_r\cdot\psi_r(t) = \boldsymbol{a'\cdot \psi}(t) \quad ; \quad \beta(t) =  \sum_{m=1}^M b_m \cdot \phi(t) = \boldsymbol{b'\cdot \phi}(t) \quad ; \quad t\in[0,T]
$$
où $\boldsymbol{\phi} = (\psi_1,\ldots,\psi_R)$ et $\boldsymbol{\phi} = (\phi_1,\ldots,\phi_M)$ sont des vecteurs de fonctions dans $L^2([0,T])$ et $\boldsymbol{a} = (a_1, \ldots ,a_R)$ et $\boldsymbol{b} = (b_1,\ldots,b_M)$ sont des points de $\mathds{R}^d$. On suppose que les fonctions $(\phi_i)_i$ et $(\psi_j)_j$ forment une base d'un sous espace vectoriel de $L^2([0,T])$. On est en train de supposer que $X$ et $\beta$ sont dans ces espaces fonctionnels de dimension finie.

Sous cette hypothèse on peut réécrire le produit scalaire $\langle X,\beta  \rangle_{L^2([0,T])}$:
$$
\langle X,\beta  \rangle_{L^2([0,T])} = \sum_{r=1}^R \sum_{m=1}^M a_r \big( \int_{[0,1]} \psi_r(s) \phi_m(s) ds \big) b_m = \boldsymbol{a'\cdot R \cdot b} .
$$
Où $\boldsymbol{R}$ est la matrice avec le produit scalaire entre les fonctions des deux bases: $$\boldsymbol{R}_{i,j} = \langle \psi_i, \phi_j\rangle_{L^2([0,T])} \quad ; \quad i\in \llbracket 1 , R \rrbracket \quad ; \quad j\in \llbracket 1 , M \rrbracket .$$ 

Si on pose $\boldsymbol{\Tilde{x} =  a'\cdot R} \in \mathds{R}^M$, alors on a: $\langle X,\beta  \rangle_{L^2([0,T])} = \langle \boldsymbol{\Tilde{x},b} \rangle_{\mathds{R}^M}$.

Ainsi le modèle  FOLR, voir (\ref{eq:FOLR}), s'écrit:
\begin{align*}
        &Y^* = \langle \boldsymbol{\Tilde{x},b} \rangle_{\mathds{R}^M} + \epsilon \stepcounter{equation}\tag{\theequation}\label{eq:FOLR_reduit},\\
        &Y = j \quad \Leftrightarrow \quad \tau_{j-1} < Y^*  \leq \tau_j .
\end{align*}
On reconnaît un modèle ordinal classique avec $M$ covariables dont les paramètres sont les seuils $\tau_1,\ldots,\tau_{K-1}$ et le vecteur $\boldsymbol{b}$ avec les coefficients de $\beta$ dans la base $(\phi_1,\ldots,\phi_M)$. Avec cette hypothèse, connaître la fonction $\beta$ est équivalent à connaître le vecteur $\boldsymbol{b}$. Ainsi, il suffit d'estimer ces coefficients de façon classique sur le modèle ordinal classique.

Notons qu'on pourrait souhaiter estimer les coefficients de $\beta$ en maximisant la vraisemblance avec un terme de pénalisation LASSO. Surtout si on utilise des fonctions $\phi_1, \ldots \phi_M$ qui ont des supports relativement distincts en $[0,T]$, comme des splines. De façon heuristique, on aurait ainsi une information sur le support de $\beta$. On l'implémenter avec des programmes qui font de l'estimation sur de modèles ordinale classique par maximum de vraisemblance avec une pénalisation LASSO, telles comme \og OrdinalNet \fg{} en R \cite{ordinalNet}.

\section{Applications}

\subsection{Données Kneading}

On propose d'illustrer les sections 2 et 3 avec le jeux de données \og Kneading \fg{}. Ce jeu de données provient du Danone Vitapole Paris Research Center et contient des mesures de résistance lors du pétrissage de 115 pâtes à cookies. Toutes les 2 secondes on mesure la résistance, cela pendant 8 minutes, et on cuit chaque pâte pour obtenir des cookies. Ensuite on va mesurer la qualité du cookie et voir s'il est \og bon \fg, \og ajustable \fg{} ou \og mauvais \fg{}. Dans cette section on essayera de prédire la qualité du cookie en utilisant la résistance de la pâte lors du pétrissage. On est clairement dans le cadre décrit dans cet article: la covariable est de nature fonctionelle et la variable cible est de nature ordinale: mauvais \textless ajustable \textless bon. 

\begin{figure}[h!]
     \centering
     \begin{subfigure}[b]{0.49\textwidth}
         \centering
         \includegraphics[width=\textwidth]{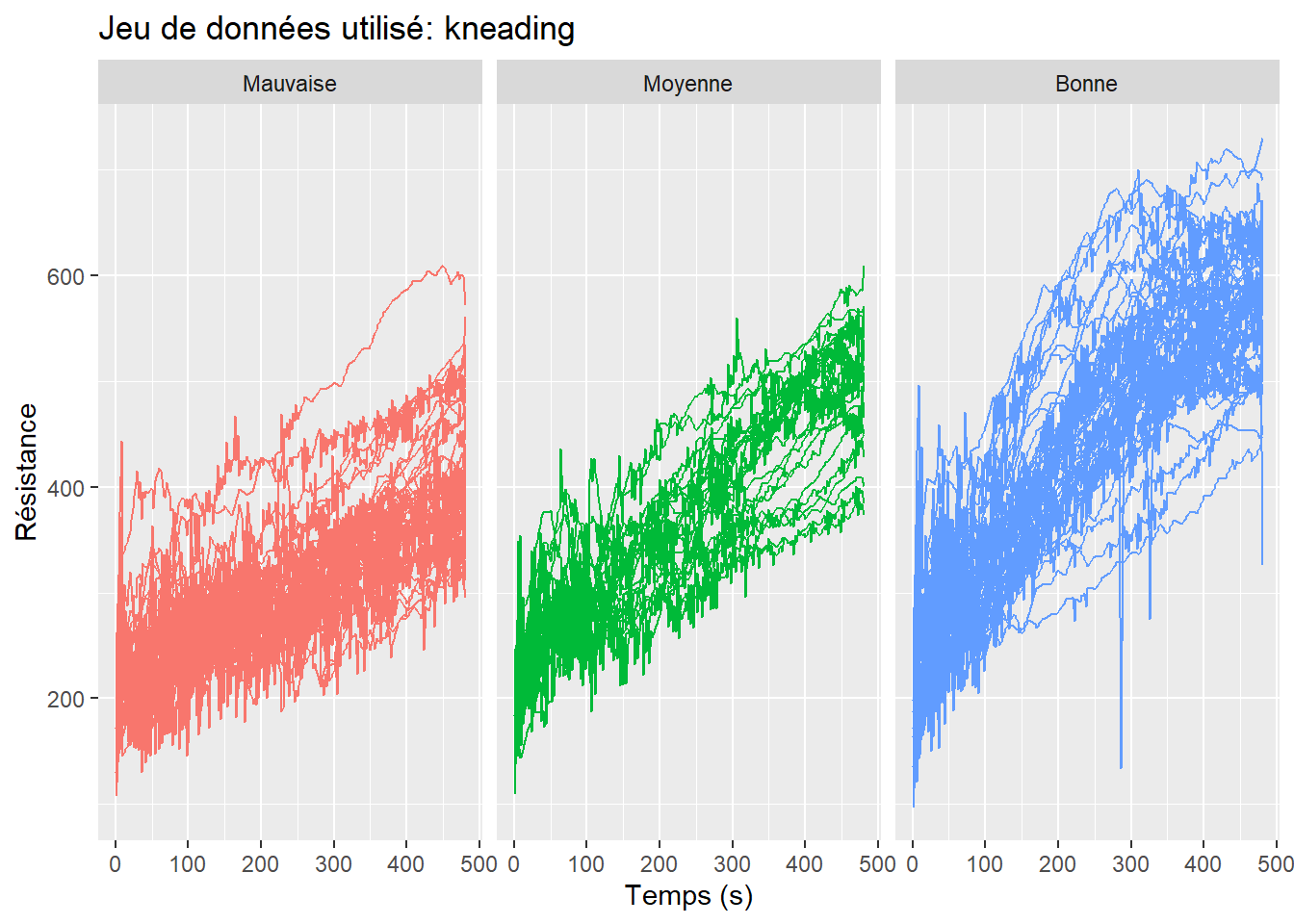}
         \caption{Trajectoires brutes}
         \label{fig:cookies_bruts}
     \end{subfigure}
     \hfill
     \begin{subfigure}[b]{0.49\textwidth}
         \centering
         \includegraphics[width=\textwidth]{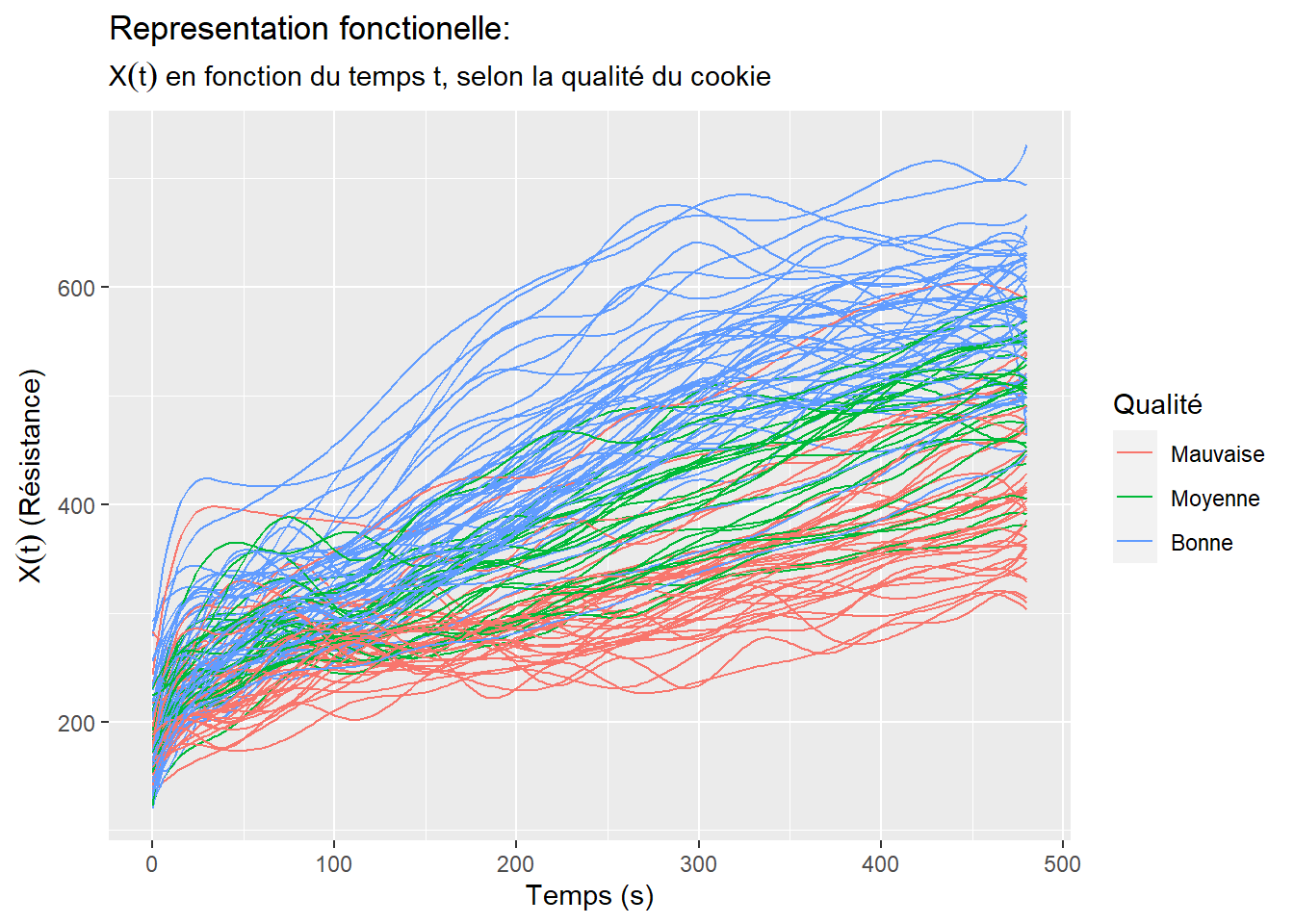}
         \caption{Trajectoires lissées}
         \label{fig:cookies_lisses}
     \end{subfigure}
        \caption{Visualisation des trajctoires selon qualité du cookie}
        \label{fig:cookies}
\end{figure}

On propose de visualiser les différentes trajectoires brutes, selon la qualité du cookie obtenue (voir la figure \ref{fig:cookies_bruts}). On s'aperçoit déjà que les bons cookies semblent avoir une résistance plus importante que les mauvais cookies, surtout vers la fin du processus.

On lisse les données brutes en les représentants de façon fonctionnelle avec une base de 16 splines d'ordre 4, en minimisant l'écart au carré. (voir la figure \ref{fig:cookies_lisses}). Comme ça on obtient les coefficients $a_1, \ldots, a_{16}$ pour chaque trajectoire.

\begin{figure}[h!]
     \centering
     \begin{subfigure}[b]{0.49\textwidth}
         \centering
         \includegraphics[width=\textwidth]{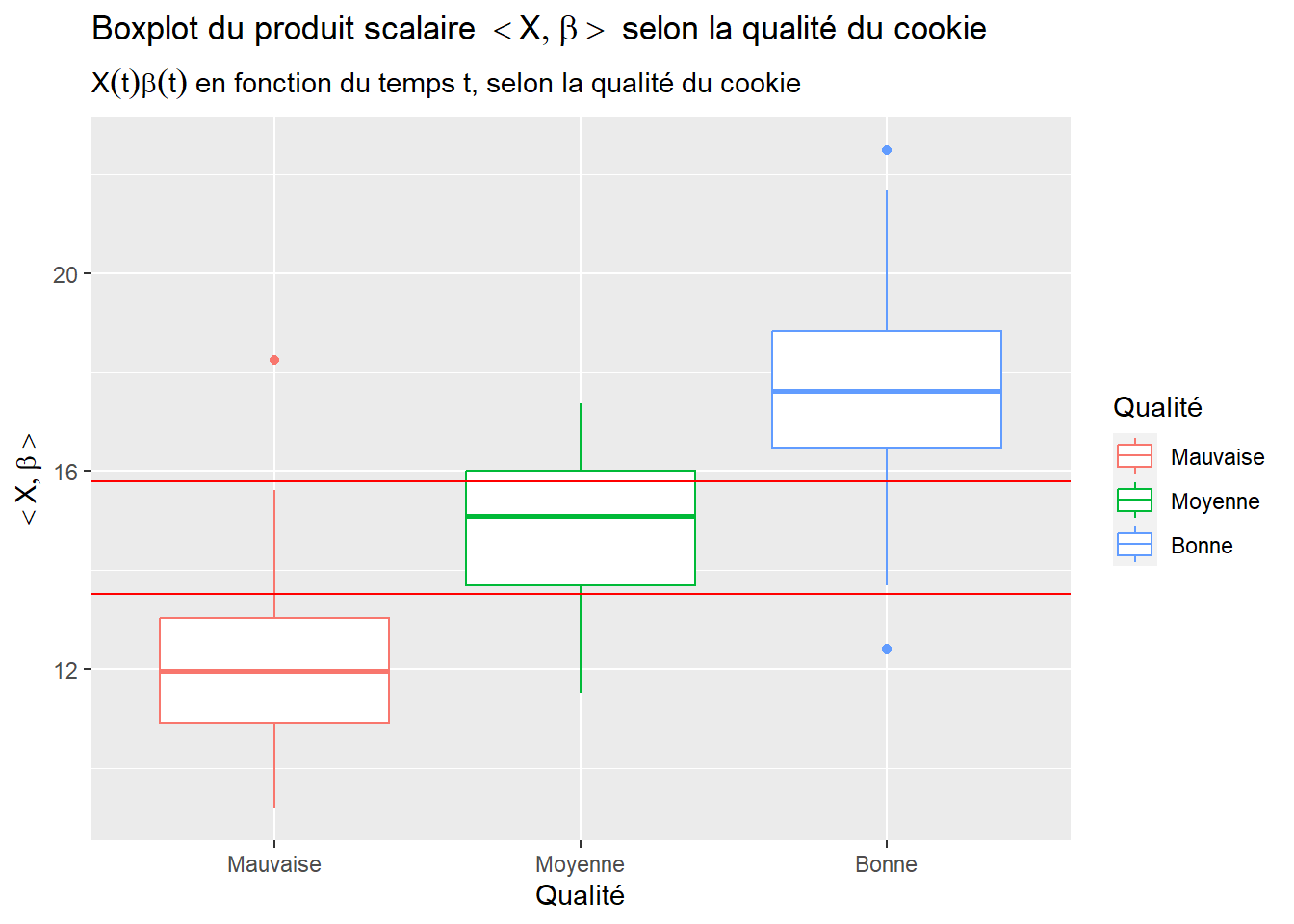}
         \caption{Produit scalaire}
         \label{fig:PS_deg2}
     \end{subfigure}
     \hfill
     \begin{subfigure}[b]{0.49\textwidth}
         \centering
         \includegraphics[width=\textwidth]{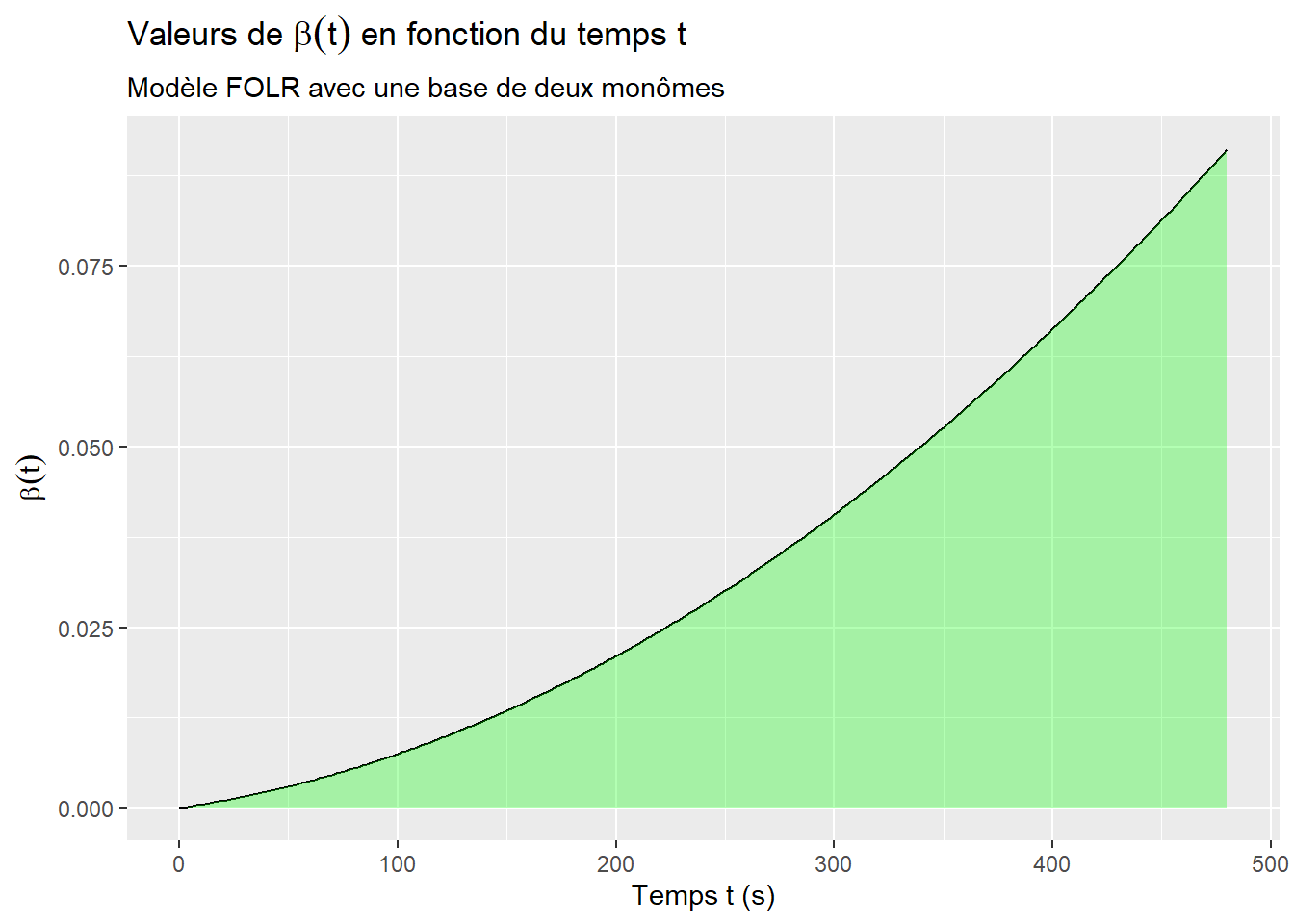}
         \caption{Fonction $\beta$ obtenue}
         \label{fig:beta}
     \end{subfigure}
        \caption{Résultats du modèle FOLR}
        \label{fig:FOLR}
\end{figure}

Avec ces coefficients là, on peut ajuster le modèle FOLR, en supposant que la fonction $\beta$ est de la forme:
$$ \beta(t)  = b_1 t + b_2 t^2 .$$
On utilise juste une base de deux monômes, on n'inclut pas de terme constant pour avoir un modèle identifiable. Le modèle donne une fonction $\beta$ et des seuils (voir les figures de \ref{fig:FOLR}). Le produit scalaire en $L^2([0,T])$ entre la fonction $\beta$ et les données fonctionnelles $X_1, \ldots , X_n$ permet de donner des prédictions optimales dans un sens, on observe bien qu'on retrouve l'ordinalité dans le produit scalaire qu'on attend (voir figure \ref{fig:PS_deg2}). Et on voit bien que la résistance vers la fin du pétrissage semble être un bon indicateur sur la qualité du cookie, c'est là où les valeurs de $\beta$ sont les plus élevées (voir figure \ref{fig:beta}).

\subsection{Données de EssilorLuxottica}

EssilorLuxottica développe des lunettes connectées équipées d'un verre E-chromic qui peut changer de teinte à la demande de l'utilisateur. Les données utilisées sont issues d'une campagne d'acquisition en extérieur à Grenoble. Il est demandé au porteur des lunettes de changer la teinte du verre à n'importe quel instant. Des mesures de lumière sont enregistrées toutes les secondes (avec un capteur ALS, \texttt{Ambient Light Sensor}) et on dispose de la teinte des verres choisie à chaque moment. On s'intéresse à comprendre et prédire la préférence du porteur en fonction de la lumière ambiante. On propose de regarder les moments où l'utilisateur a changé de teinte et prédire la nouvelle teinte qu'il choisira.

On comparera le modèle ordinal avec une covariable fonctionnelle (les mesures ALS dans une fenêtre de temps) au modèle ordinal classique qui utilise juste la toute dernière mesure de l'ALS avant le changement de teinte.

\begin{figure}[]
     \centering
     \begin{subfigure}[b]{0.49\textwidth}
         \centering
         \includegraphics[width=\textwidth]{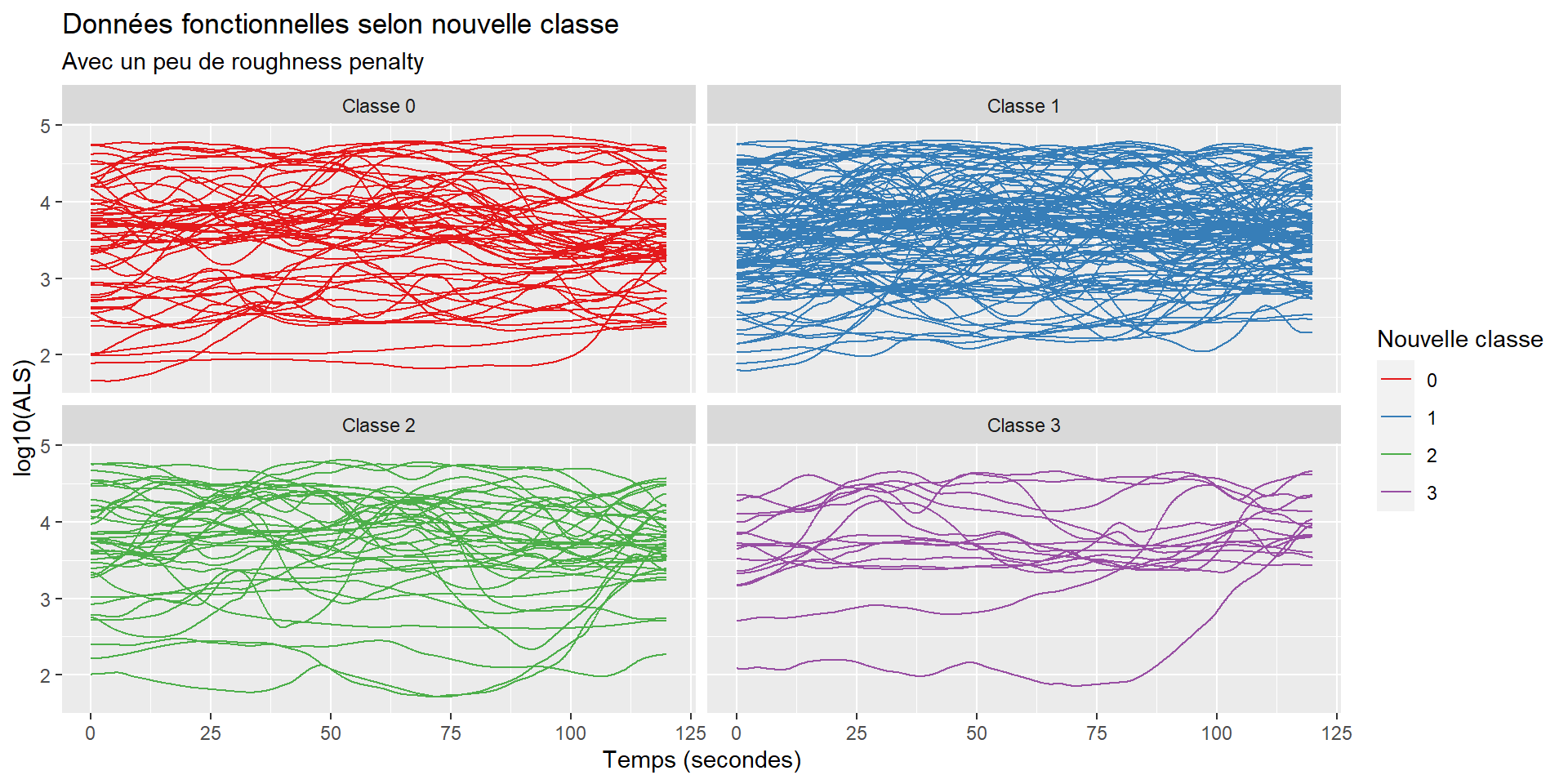}
         \caption{Observations selon nouvelle teinte}
         \label{fig:fd_brut}
     \end{subfigure}
     \hfill
     \begin{subfigure}[b]{0.49\textwidth}
         \centering
         \includegraphics[width=\textwidth]{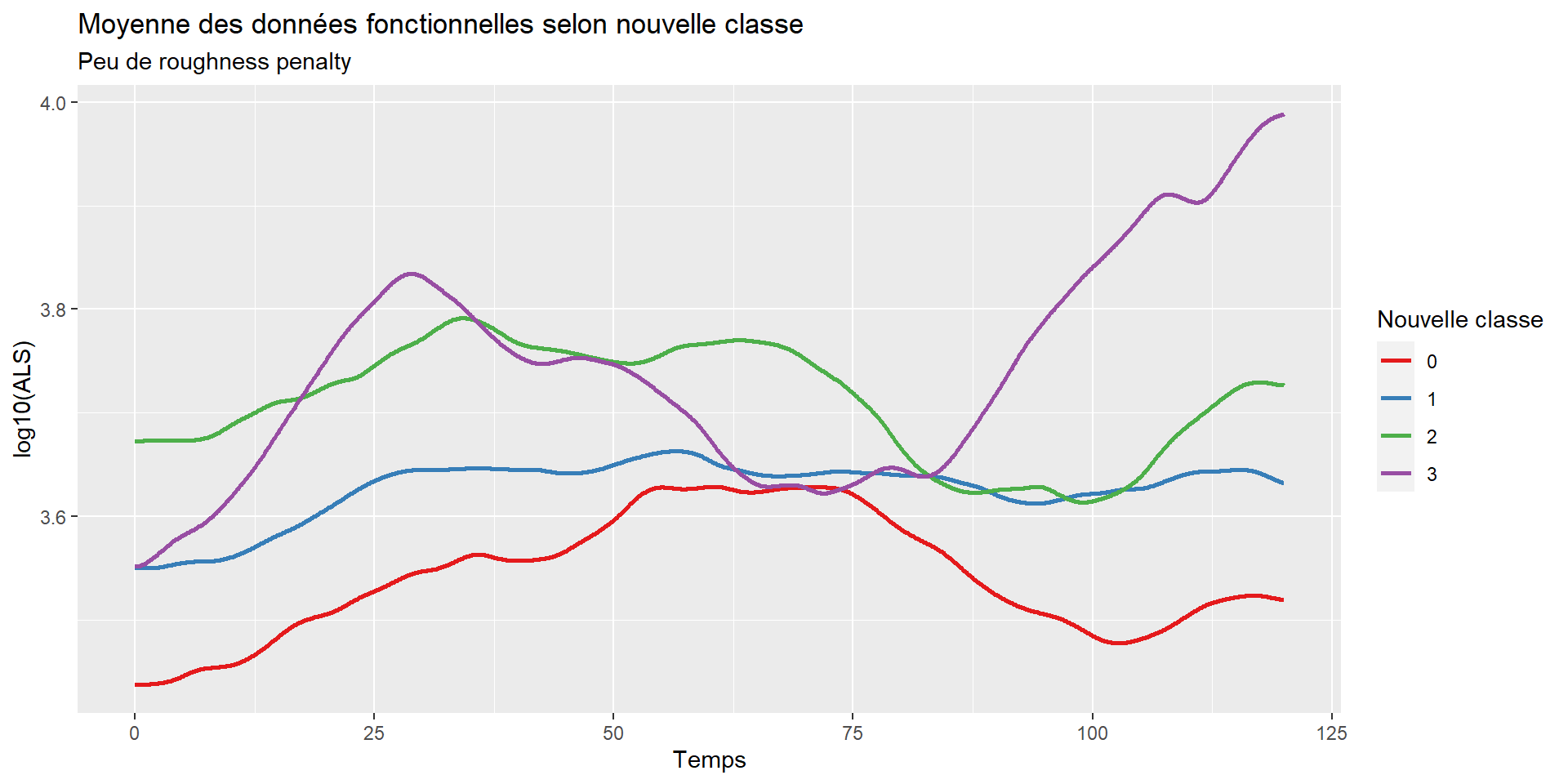}
         \caption{Moyennes selon nouvelle teinte}
         \label{fig:fd_moyen}
     \end{subfigure}
        \caption{Données fonctionnelles}
        \label{fig:fd}
\end{figure}

On garde 233 moments de changement de teinte, on regardera les deux dernières minutes avant le changement de classe. En utilisant une base de 60 splines, avec une \og roughness penalty \fg{} de $10^{-0.5}$ on obtient les données fonctionnelles de la figure \ref{fig:fd_brut}.

On ne note pas de claire différence entre les classes en regardant directement les données fonctionnelles, par contre en observant les moyennes selon la nouvelle classe (voir figure \ref{fig:fd_moyen} ) on note que l'on retrouve une certaine ordinalité dans ces données: juste avant le changement de teinte, avoir une haute mesure de luminosité indique que l'utilisateur changera vers une teinte obscure, inversement une faible mesure de luminosité indique qu'il choisira une  teinte claire. Puis les moyennes vers la fin de la fenêtre temporelle sont ordonnées de la classe plus claire (la classe 0) à la plus obscure (classe 3).

\begin{figure}[h!]
     \centering
     \begin{subfigure}[b]{0.49\textwidth}
         \centering
         \includegraphics[width=\textwidth]{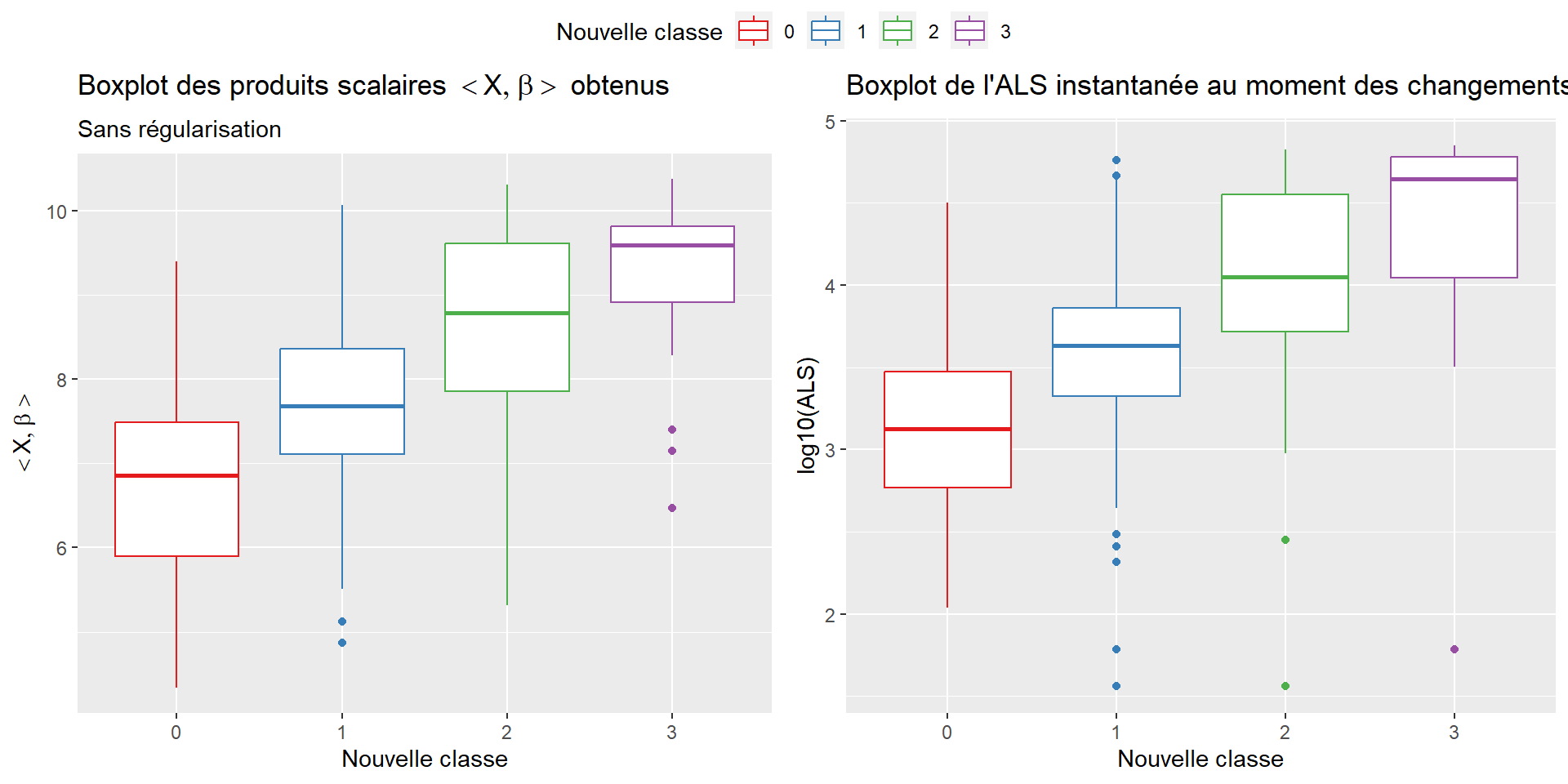}
         \caption{Produit scalaire}
         \label{fig:FOLR_gren_ps}
     \end{subfigure}
     \hfill
     \begin{subfigure}[b]{0.49\textwidth}
         \centering
         \includegraphics[width=\textwidth]{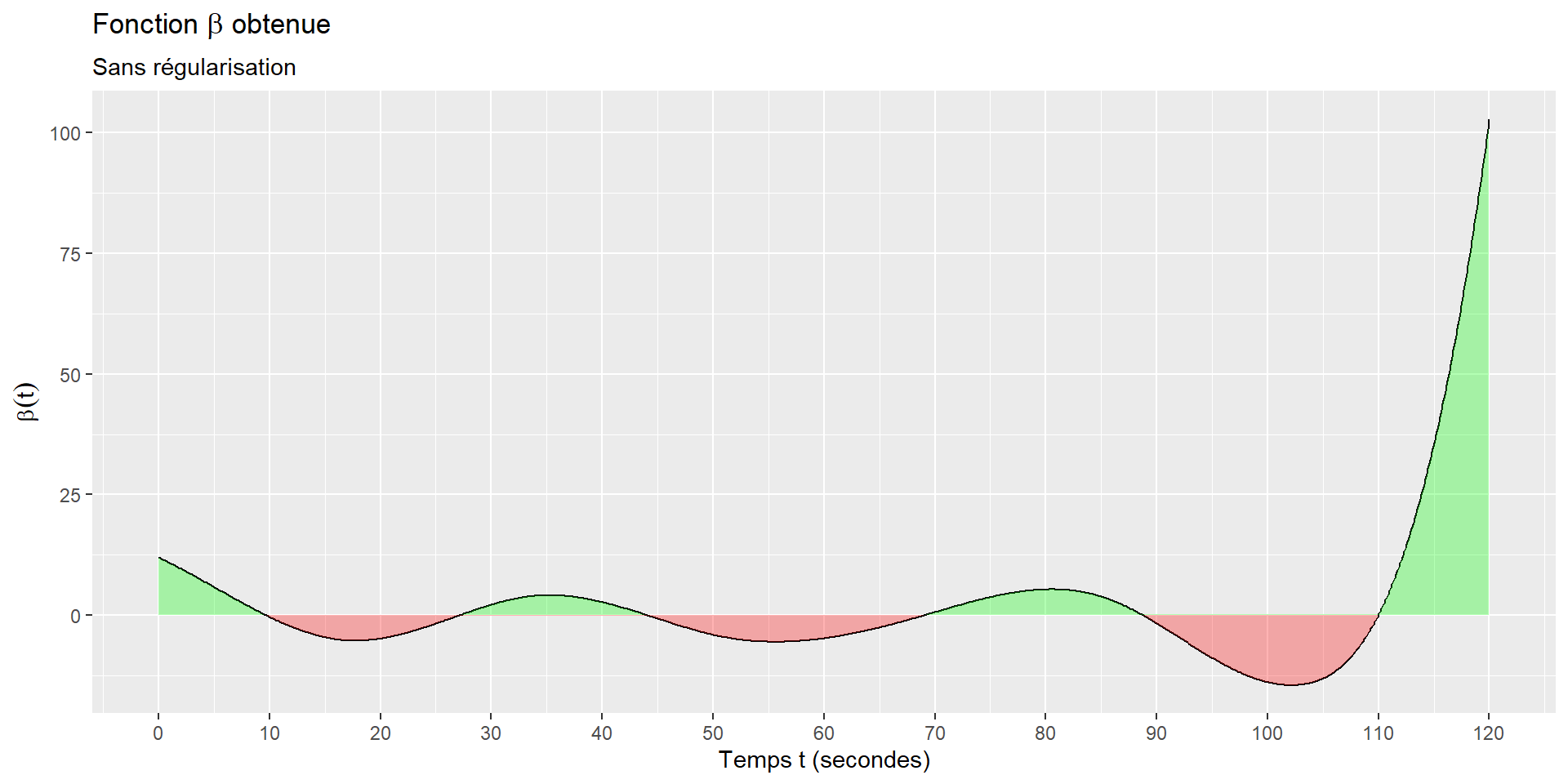}
         \caption{Fonction $\beta$ obtenue}
         \label{fig:FOLR_gren_beta}
     \end{subfigure}
        \caption{Modèle FOLR: données d'EssilorLuxottica}
        \label{fig:FOLR}
\end{figure}

Dans un premier temps, on ajuste un modèle FOLR sur les données fonctionnelles, en prenant une base de 10 splines pour $\beta$. Ainsi le modèle permet de donner une estimation de la fonction $\beta$ (voir figure \ref{fig:FOLR_gren_beta}), des seuils $\tau_1 , \tau_2 , \tau_3$ ordonnés et on peut regarder le produit scalaire $\langle X_i, \beta \rangle_{L^2([0,T])}$ avec $i \in \llbracket 1,233 \rrbracket$ (voir figure \ref{fig:FOLR_gren_ps}) . Ce dernier devrait être petit pour les teintes claires et grand pour les classes obscures, vu que l'on obtient une fonction $\beta$ qui est surtout positive, c'est bien ce que l'on observe mais ce n'est pas clair que le modèle FOLR discrimine mieux les futures classes que le modèle ordinal simple (voir la figure \ref{fig:FOLR_gren_ps}), on remarque aussi que la fonction $\beta$ oscille avec des faibles valeurs avant la seconde 90 et prend une grande valeur vers la fin. Il se pourrait que $\beta$ soit nulle avant la seconde 90 et le modèle suggère que la dernière observation a une forte importance sur la nouvelle classe choisie.

\begin{figure}[h!]
     \centering
     \begin{subfigure}[b]{0.49\textwidth}
         \centering
         \includegraphics[width=\textwidth]{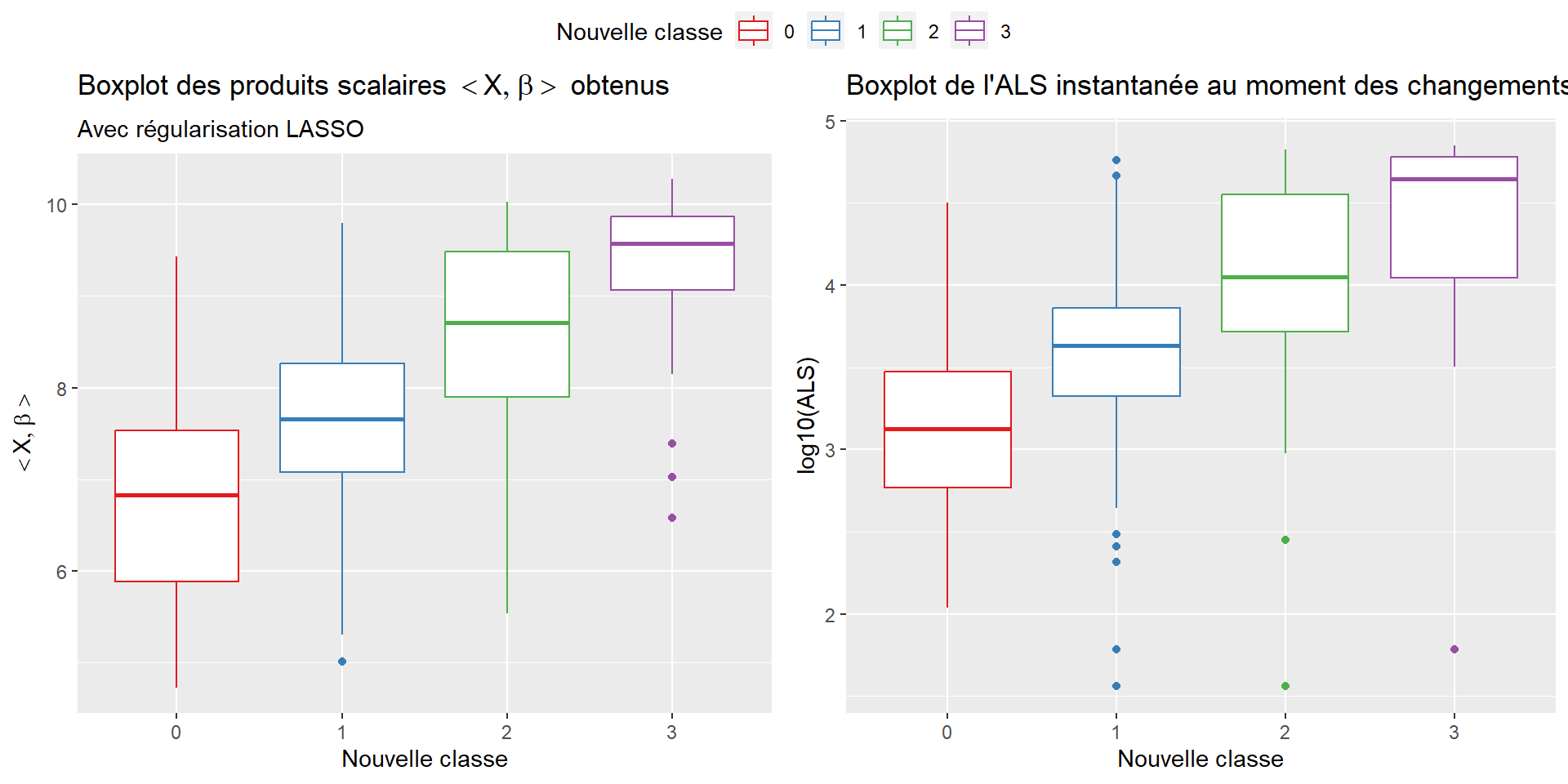}
         \caption{Produit scalaire}
         \label{fig:FOLR_gren_lasso_ps}
     \end{subfigure}
     \hfill
     \begin{subfigure}[b]{0.49\textwidth}
         \centering
         \includegraphics[width=\textwidth]{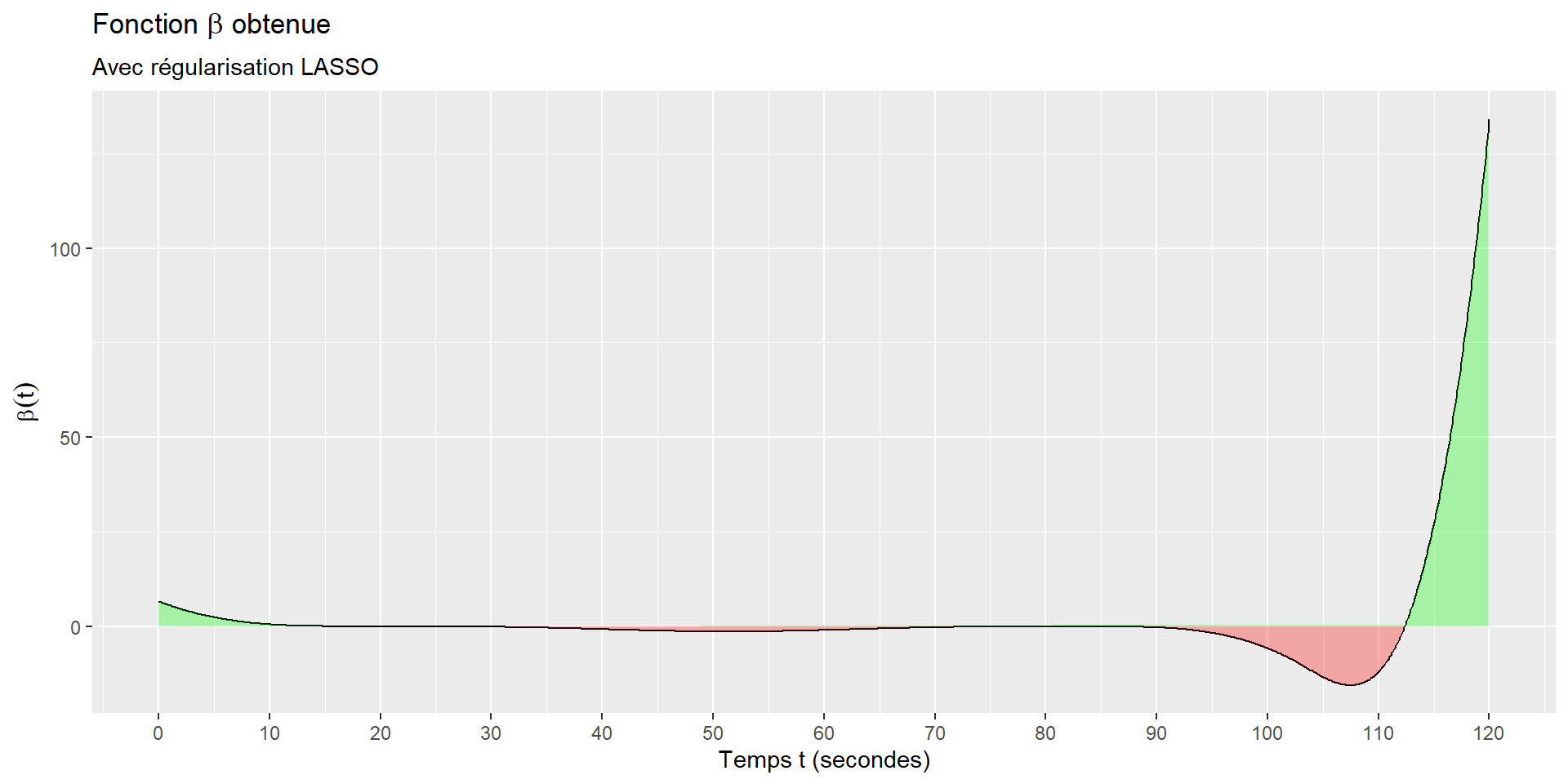}
         \caption{Fonction $\beta$ obtenue}
         \label{fig:FOLR_gren_lasso_beta}
     \end{subfigure}
        \caption{Modèle FOLR avec pénalisation LASSO: données d'EssilorLuxottica}
        \label{fig:FOLR}
\end{figure}

Dans un deuxième temps, on reprend le modèle FOLR et on utilise une pénalisation LASSO, on obtient $\beta_{LASSO}$ (voir figure \ref{fig:FOLR_gren_lasso_beta}) et des produits scalaires $\langle X_i, \beta_{LASSO} \rangle_{L^2([0,T])}$ avec $i \in \llbracket 1,233 \rrbracket$ (voir figure \ref{fig:FOLR_gren_lasso_ps}). On remarque que $\beta_{LASSO}$ est nulle avant la seconde 90 et après, il prend une forte valeur vers la dernière valeur. Ainsi, ce dernier modèle suggère que les seules observations qui auront un impact sur la prédictions sont celles qui arrivent les dernières 30 secondes et le moment le plus important pour la prédiction est le dernier. Les valeurs négatives de $\beta_{LASSO}$ pourraient être un artefact de l'estimation, vu qu'on utilise des splines pour représenter $\beta_{LASSO}$, la vraie fonction pourrait ne pas être très régulière et les splines le sont.

On comparera les prédictions des 3 modèles : régression ordinale en utilisant la dernière valeur de l'ALS, les régressions FOLR sur la série des mesures ALS des deux dernières minutes, avec ou sans pénalisation LASSO. Pour cela on utilisera un \og 10-fold \fg{} sur les données et on utilisera l'erreur de précision et l'erreur en valeur absolue moyen comme métriques (en prenant la moyenne aussi sur les 10 \og folds \fg{}). On obtient les résultats de la table \ref{tab:results}. 

Les résultats obtenus ne sont pas concluants, ce modèle à covariables fonctionnelles (avec ou sans pénalisation LASSO) n'est pas plus performant que celui qui prend la dernière valeur ALS, et aucun des modèles donne une prédiction satisfaisante. Cela dit, un possible avantage d'utiliser le modèle FOLR est qu'il prend une moyenne pondérée dans le temps, on s'attend à qu'il soit moins sensible à des courtes et fortes variations de luminosité.
\begin{center}

\begin{longtable}{lrr}
\caption{
{\large Erreurs de generalisation des predictions} \\ 
{\small avec un 10-fold}
}\label{tab:results} \\ 
\toprule
Prédiction & Erreur en valeur absolue & Erreur de précision \\ 
\midrule
ALS instantanee & $0.48$ & $44\%$ \\ 
FOLR sans penalisation & $0.53$ & $48\%$ \\ 
FOLR avec penalisation LASSO & $0.50$ & $45\%$ \\ 
\bottomrule
\end{longtable}
\end{center}
\section*{Conclusion}

On a proposé une solution au problème industriel, le modèle ordinal à covariables fonctionnelles devrait pouvoir donner une réponse qui prend en compte la nature fonctionnelle des données et la nature ordinale des classes de teinte. 

Hors, un point important qu'on ne traite pas dans cet article est le choix des bases fonctionnelles utilisées, une fois que ce choix est fait, on a montré que le problème fonctionnel devient un problème multivarié classique qui est \og agnostique \fg{} de la base utilisée. Ainsi ce choix est important et il n'est pas justifié formellement dans ce document. Si la prédiction faite sur les données d'EssilorLuxoticca n'est pas satisfaisante, c'est peut-être lié au fait que la base de splines ne serait pas bien adaptée à un signal très irrégulier comme celui de l'ALS. 

En outre, dans la pratique, on n'observe pas des fonctions, mais un nombre fini de mesures à certains instants, en plus de cela ces mesures peuvent être bruitées, donc il y a en réalité une première estimation qui est faite avant d'ajuster le modèle. On pourrait envisager d'étudier théoriquement comment faire l'estimation jointe des coefficients du signal et ceux du modèle, cela pourrait permettre de justifier le choix d'une famille de bases plutôt qu'une autre.

\bibliography{biblio}

\begin{thebibliography}{}

\bibitem[Agresti, 2010]{agresti_categorical}
Agresti, A. (2010).
\newblock {\em Analysis of Ordinal Categorical Data}.
\newblock Wiley Series in Probability and Statistics. Wiley.

\bibitem[Hastie et~al., 2009]{stat_learning}
Hastie, T.~J., Tibshirani, R.~J., and Friedman, J.~H. (2009).
\newblock {\em The elements of statistical learning: data mining, inference,
  and prediction.}
\newblock Springer series in statistics. Springer.

\bibitem[Jacques and Samard{\v z}i{\'c}, 2022]{regOrd_fonctionelle}
Jacques, J. and Samard{\v z}i{\'c}, S. (2022).
\newblock {Analyzing cycling sensors data through ordinal logistic regression
  with functional covariates}.
\newblock {\em {Journal of the Royal Statistical Society: Series C Applied
  Statistics}}.

\bibitem[M{\"u}ller and Stadtm{\"u}ller, 2005]{mouches}
M{\"u}ller, H.-G. and Stadtm{\"u}ller, U. (2005).
\newblock Generalized functional linear models.
\newblock {\em The Annals of Statistics}, 33(2):774 -- 805.

\bibitem[Preda et~al., 2010]{cookies}
Preda, C., Saporta, G., and Mbarek, M.~H. (2010).
\newblock Anticipated and adaptive prediction in functional discriminant
  analysis.
\newblock In Lechevallier, Y. and Saporta, G., editors, {\em Proceedings of
  COMPSTAT'2010}, pages 189--198, Heidelberg. Physica-Verlag HD.

\bibitem[Ramsay and Silverman, 2005]{fda_ramsay}
Ramsay, J.~O. and Silverman, B.~W. (2005).
\newblock {\em Functional data analysis. [Texte imprimé].}
\newblock Springer series in statistics. Springer.

\bibitem[Wurm et~al., 2021]{ordinalNet}
Wurm, M.~J., Rathouz, P.~J., and Hanlon, B.~M. (2021).
\newblock Regularized ordinal regression and the ordinalnet r package.
\newblock {\em Journal of Statistical Software}, 99(6):1--42.

\end{thebibliography}

\newpage
\appendix

\end{document}